\documentclass[conference]{IEEEtran}
\IEEEoverridecommandlockouts
\usepackage{cite}
\usepackage{amsmath,amssymb,amsfonts}
\usepackage{algorithmic}
\usepackage{graphicx}
\usepackage{textcomp}
\usepackage{subcaption}
\usepackage[table,xcdraw]{xcolor}

\def\BibTeX{{\rm B\kern-.05em{\sc i\kern-.025em b}\kern-.08em
    T\kern-.1667em\lower.7ex\hbox{E}\kern-.125emX}}

\makeatletter
\newcommand{\linebreakand}{%
  \end{@IEEEauthorhalign}
  \hfill\mbox{}\par
  \mbox{}\hfill\begin{@IEEEauthorhalign}
}
\makeatother

\begin{document}

\title{Towards hardware Implementation of WTA for CPG-based control of a Spiking Robotic Arm\\
\thanks{This research was partially supported by the Spanish grants MINDROB (PID2019-105556GB-C33/AEI/10.13039/501100011033), by the SMALL (PCI2019-111841-2/AEI/10.1309/501100011033) and by the European Union’s Horizon 2020 ERC project NeuroAgents under Grant 724295. E. P.-F. was supported by a "Formaci\'{o}n de Personal Universitario" Scholarship from the Spanish Ministry of Education, Culture and Sport.}
}

 \author{
 \IEEEauthorblockN{A. Linares-Barranco, E. Piñero-Fuentes, \\ S. Canas-Moreno, A. Rios-Navarro}
 \IEEEauthorblockA{\textit{Robotics and Technology of Computers Lab. I3US. SCORE. } \\
 \textit{University of Seville}. Seville, Spain \\
 alinares@atc.us.es}%https://orcid.org/0000-0002-6035-9010,
 \and
 \IEEEauthorblockN{Maryada, Chenxi Wu, Jingyue Zhao,\\ D. Zendrikov, G. Indiveri}
 \IEEEauthorblockA{\textit{Institute of Neuroinformatics} \\
 \textit{University of Zurich and ETH Zurich}\\
 Zurich, Switzerland}
 }

\maketitle

\begin{abstract}
Biological nervous systems typically perform the control of numerous degrees of freedom for example in animal limbs. Neuromorphic engineers study these systems by emulating them in hardware for a deeper understanding and its possible application to solve complex problems in engineering and robotics. Central-Pattern-Generators (CPGs) are part of neuro-controllers, typically used at their last steps to produce rhythmic patterns for limbs movement. Different patterns and gaits typically compete through winner-take-all (WTA) circuits to produce the right movements. In this work we present a WTA circuit implemented in a Spiking-Neural-Network (SNN) processor to produce such patterns for controlling a robotic arm in real-time. The robot uses spike-based proportional-integrative-derivative (SPID) controllers to keep a commanded joint position from the winner population of neurons of the WTA circuit. Experiments demonstrate the feasibility of robotic control with spiking circuits following brain-inspiration. 
\end{abstract}


\section{Introduction}
Understanding how neural circuits in animal brains carry out sensory processing, decision making, and motor control can lead to useful engineering solutions for robotics and computing.
Insects are particularly interesting in this research area due to their relatively low complexity when compared to mammals' brains.
The neural circuits controlling insect limbs are individually modified and fine-tuned in a task-dependent fashion~\cite{BUSCHGES2012602}.
The output stage of these circuits results in specific motor actions for a particular degree-of-freedom (DoF) of the limb.
Conversely, circuits in the previous processing layers implement many other computational primitives, such as central-pattern-generators (CPGs) for the generation of repetitive sequences of motor actions, coordinated among the different DoFs of multiple limbs~\cite{Donati2016,Krause_etal21}, or Winner-Take-All (WTA) networks for selective attention and decision making.
% Muscles in mammals are controlled in a more complex way, including typically three levels of abstraction for the motor system: motor behaviour, limb mechanics and neural control, which are managed by the spinal cord, brainsteam and cerebral cortex.
% The neural control in this case is demonstrated not to be a copy from one arm / leg to another symetric one when performing the same task \cite{Pellegrino2020}, what probes that neural control is learnt independently instead of obeying a pre-programmed pattern.

Although there is strong evidence from neuroscience that there are well defined motor primitives for biological movement control~\cite{Stroud_etal18}, the neural basis of many of most of these primitives is still poorly understood.
% In~\cite{Hart1322} multielectrode recordings of the spianl cord gray of spinalized frogs were conducted to infer premotor drive patterns, and they observed that sets of  spiking neurons configured as a hard-WTA: that is, a network in which only one excitatory cluster with the highest spiking activity wins, while all others are suppressed.

One promising hypothesis is that the activation of sequences of patterns comes from motor primitives located in high level layers in the motor cortex, where decision making circuits are driven by sensory processing ones, and generate dynamic activities that produce motor behaviour \cite{Latorre2013}. Given these sets of complex dynamic transformations, it is necessary to model neural systems using a non-linear dynamics approach \cite{Rabinovich_etal06}. Following this approach, we propose to develop motor primitives based on WTA neural circuits, for producing the final sequential pattern of motor commands to be applied to the limb.
Examples of neuromorphic electronic systems that comprise multiple coupled WTA networks of silicon (spiking) neurons have been proposed, that give rise to noise-robust patterns of sequential activity to produce robust patterns that can be halted, resumed and readily modulated by external input \cite{Neftci_etal13}, \cite{Mostafa_etal14}. Neuromorphic mixed-signal circuits help in validating computational neuroscience models (e.g., of motor primitives) by directly emulating their signal processing features and dynamics via the properties of their analog circuits comprising transistors that operate in the weak-inversion regime \cite{Chicca_etal14}. An example of this approach was recently proposed, in which a a close-loop motor controller was developed using a WTA, implemented with silicon neurons present on the DYNAP-SE mixed-signal neuromorphic processor \cite{Moradi_etal18} and interfaced to a humanoid robot \cite{Zhao_etal20}. A second, alternative, approach is to validate the different types of motor controllers in closed-loop setups by simulating the SNN using digital SNN chips, such as the Intel Loihi or SpiNNaker ones \cite{Lele2021}\cite{Stagsted_etal20}\cite{Tieck2018}.
In \cite{Lele2021} similar approaches have been developed to manage a real robot, where the weights of a bio-mimetic multigait CPG have been programmed and coupled with DVS based visual data processing to show a spike-only closed-loop robotic system for a prey-tracking scenario for a locomotion experiment in an hexapod robot. The speed and direction of each leg is commanded through the SCPG spikes implemented in a software running in an embedded computer, which are mapped to produce the gaits through pulse-width-modulation (PWM) signals sent to the servo motors. In \cite{edbiorob2020} the final actuation of the motors of the robotic arm joints is done also in the spiking domain through the pulse-frequency-modulation (PFM), allowing lower power and latency with respect to classic robotics. This PFM was proposed in \cite{spid_sensors_2012} and later
used by \cite{Donati_etal18} for open-loop control. It is currently available at the ED-Biorob \cite{edbiorob2020} and the ED-Scorbot framework, which can be used remotely if needed \cite{EDScorbot-Springer}. For the best of our knowledge the first work that presented a spiking interface from an SNN hardware accelerator and a PFM robotic platform is ED-Biorob. Nevertheless, in this work, authors required the use of all the neurons a chip (1Kneurons) to produce the needed spiking activity to make a joint to reach the farthest position. This restricts the use of the silicon neurons of the chip for implementing an SNN for controlling the robot.

In this work we use a mixed signal neuromorphic processor to implement multiple coupled WTA networks to produce and control end-effector trajectories of an event-driven robotic arm. The event-driven robotic platform ED-Scorbot has been connected to the mixed-signal neuromorphic processor where several populations of silicon neurons compete to control the joint of the robotic arm to reach a desired angle. The main difference with ED-Biorob is that in this case we require less neurons per joint, allowing the use of remain neurons to implement a SNN for the robot commanding. The address-event-representation (AER) protocol is used to codify and communicate neural activity from the neuromorphic processor chip to the SPID controllers running in parallel for the 6 DoF of this robot. This paper is focused in the interface that requires, as in the biological neural systems, to extract the right pattern from the winner population and map it to the joint movement in a smooth and safe way for the robot parts. Experiments on one joint of the robot validate the use of this architecture and open the door for future control of the whole robot in a neuro-inspired way.

%The main contribution of this paper is to reproduce with spiking hardware a WTA circuit able to control a joint of a robotic platform taking inspiration from the nervous system for a better understanding of its benefits and as a contribution for future multi-joint neuromorphic robotic controllers.

\section{Materials and methods}

\subsection{WTA network}
The Winner-Take-All (WTA) network is among most extensively studied competition models in computational neuroscience. These networks are composed of clusters of excitatory neural populations, competing via shared inhibition, and in which each cluster encodes one of the possible input features. For the purpose of this work we consider as input feature space, the space of possible angles for the robotic arm joint movements. We implemented as WTA a network of spiking neurons configures as a hard-WTA: that is, a network in which only one excitatory cluster with the highest spiking activity wins, while all others are suppressed.
The population coding strategy, and the use of clusters of neurons to represent one WTA node, allows to overcome the variability and noise present in the neuromorphic hardware.
In addition, we implement strong recurrent excitatory connections within a cluster to facilitate self-sustained activity as a means to increase robustness to noise in the input spikes.
The suppression of a cluster's activity occurs when inhibition intervenes with prevailing excitation such that the network transits from the currently active cluster to another one winner, that represents the subsequent angle movement.

\begin{figure}
 \centering
 \includegraphics[width=0.5\textwidth]{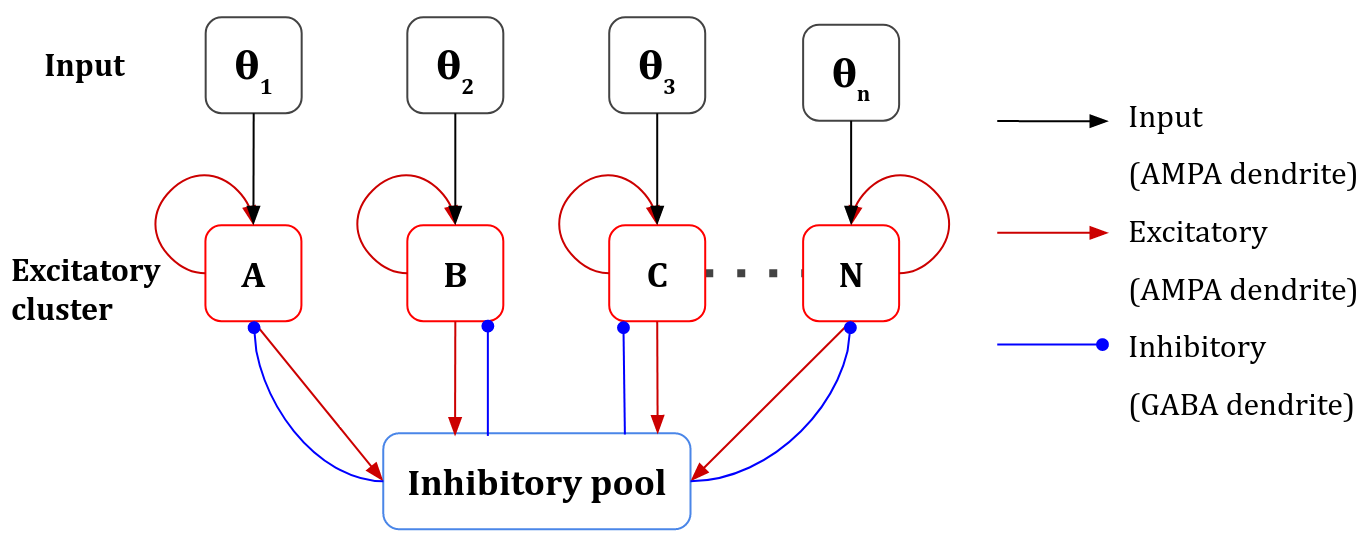}
 \caption{Winner-take-all network architecture}
 \label{fig:wta_abstract}
\end{figure}

The specific network architecture consists of 12 clusters with $N_e = 8$ excitatory neurons tuned to a specific angle.
The inhibitory pool interacts with all these clusters in such a way as to permit at-most one cluster to be active, encoding the most recent input (see Fig.~\ref{fig:wta_abstract}).
Depending on the coupling of excitatory clusters and inhibitory pool, when a new input signal arrives that activates cluster \emph{$n_{new}$} (say cluster B, in Fig.~\ref{fig:wta_abstract}), the inhibitory mechanism kills the activation of cluster \emph{$n_{prev}$} (e.g., cluster A) and concurrently balances the activity in the newly activated cluster maintaining stability.
%The contrast of excitation and inhibition is attained by AMPA and GABA dendrites (see section \ref{subsec:dynpase}).

\subsection{Infrastructure}
\subsubsection{The mixed-signal neuromorphic processor}
\label{subsec:dynpase}
To test  the network and robotic setup we use a mixed-signal neuromorphic processor of the type described in~\cite{Moradi_etal18} that uses analog circuits for the synapse and neural dynamics and digital circuits for the event-based communication.
Combining this with weak inversion aVLSI design methods (sub-threshold design) provides the chip with impeccable energy efficiency.
% It's clock free design runs in native real-time and the asynchronous communication scheme ensures very low latencies.
Each chip has 1024 exponential Integrate and Fire neurons distributed over 4 individually configurable Neural Cores, connected by hierarchical routing grid.
%remove%
% The tag-based routing infrastructure provides direct communication from one chip to 15 × 15 surrounding chips.
% So for a fully connected system a 2D grid of 8 × 8 is supported by the chips without mapping, or placing them in a sphere a total of 15 × 15 is possible without extra mapping.
%remove%
% The heart of the DynapSE2 is the exponential integrate and fire somatic model.
Each physical neuron is composed of Somatic, Dendritic and Synaptic blocks that runs in real-time and emulates complex neuron behaviour.
The Synaptic block comprises 64 content addressable memory (CAM)-based synapses per neuron, with 4-bit weights.
The Somatic block models an exponential or linear Integrate and Fire (IAF) neuron, with optional firing-rate adaptation.
The Dendritic block contains 2 excitatory and 2 inhibitory differential pair integrator-based low-pass filter compartments, which elicit excitatory and inhibitory post synaptic potentials (EPSP / IPSP).
Each synapse can be attached to any of the four dendritic compartments.
The AMPA dendritic compartment offers an optional 1D or 2D resistive grid, to diffuse incoming EPSPs between nearby somas.
The NMDA dendritic compartment can gate the incoming current depending on the somatic membrane potential.
The two GABA dendritic compartment mimics substractive and divisive inhibition.
% Both the AMPA and NDMA excitatory dendritic compartments, as well as the GABA inhibitory compartment, can be individually switched to conductance mode, to emulate a large class of biologically inspired synaptic models.

% This chip is mounted in a stackable printed-circuit-board (PCB) connected to an Opal-Kelly FPGA-based computer interface for configuring the SNN to run on the chip and for sending and extracting spiking traffic.
% A Python-based software tool allows to program and configure the SNN on the chip in a user-friendly manner. The neuron spiking output activity can be sent off-chip to another system by simply selecting the right output port from the system's four ports.
% These ports are address-event-representation (AER) ports of 15-bits, including the two lines dedicated to the asynchronous handshake protocol.
% For each output spike produced in the chip, a couple of two consecutive AER words is sent.
A software tool implementing AER communication protocol in C++ allows to program and configure a SNN, stimulate silicon neurons, and stream events out in Python. A periphery FPGA holds configurable spike generators to produce either custom or Poisson spike trains. Output spikes can either be directed to PC via the FPGA or to another AER-based system, like in this work by simply selecting the right output port from the four ports that include the on-chip router. These are AER ports of 15-bit including two lines for a simple asynchronous handshake protocol. For each output spike produced in the chip, a couple of two consecutive AER words is sent to the interfaced AER system.
This output activity can be either received by an FPGA-based interface, or by another AER system.
% When sent to the Opal-Kelly interface, Samna software can store or represent this output activity in the computer.

\subsubsection{Event-Driven Scorbot}
The Event-Driven Scorbot (ED-Scorbot) platform consists of a hardware/software infrastructure that includes three main parts.
First, a six DoF mechanical industrial robotic arm.
Second, an implementation of a SPID controller for each DoF, as in~\cite{edbiorob2020}~\cite{EDScorbot-Springer}, that is able to command a joint of the robot to move to a fixed position using spiking information~\cite{spid_sensors_2012}.
These controllers implementation is comprised of several boards, which together allow for the six parallel SPID controllers to command the whole robot.
Third, a software application-programmer-interface (API) developed in Python \cite{pyedscorbot} that allows to interface of the robot with software based spiking systems or hardware based ones, like DYNAP-SE, as in this work.

\begin{figure}
 \centering
 \includegraphics[width=0.36\textwidth]{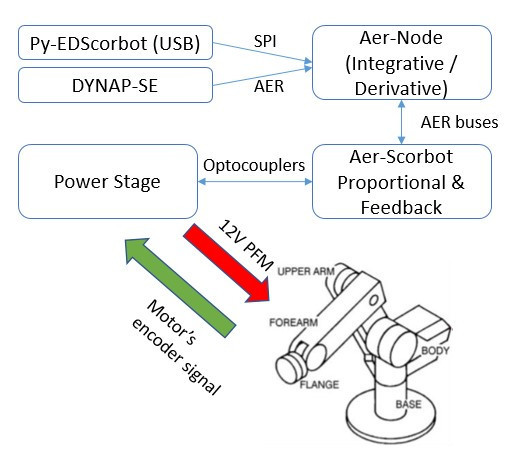}
 \caption{High-level ED-Scorbot with Dynap-se logical overview}
 \label{fig:scorbot_logical_overview}
\end{figure}

\begin{figure*}
  \begin{subfigure}{0.60\textwidth}
    \includegraphics[width=\textwidth]{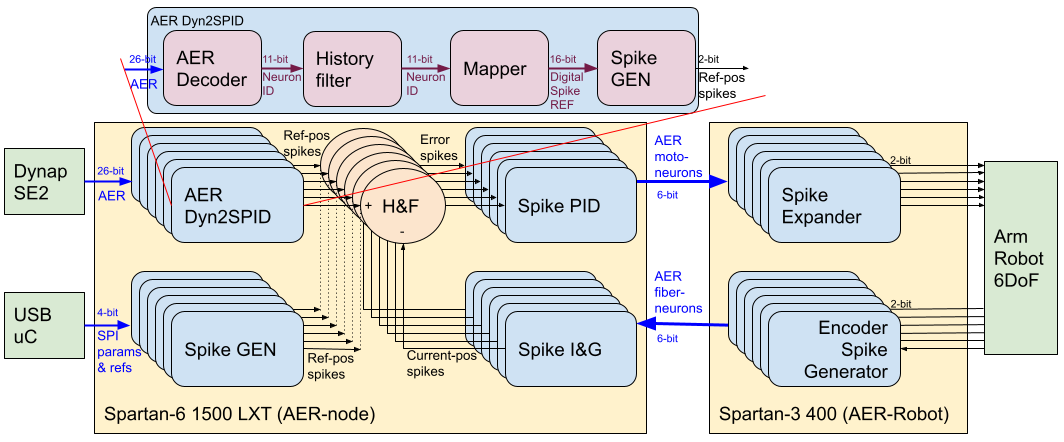}
    \subcaption{}
  \end{subfigure}
  \begin{subfigure}{0.34\textwidth}
    \includegraphics[width=\linewidth, keepaspectratio]{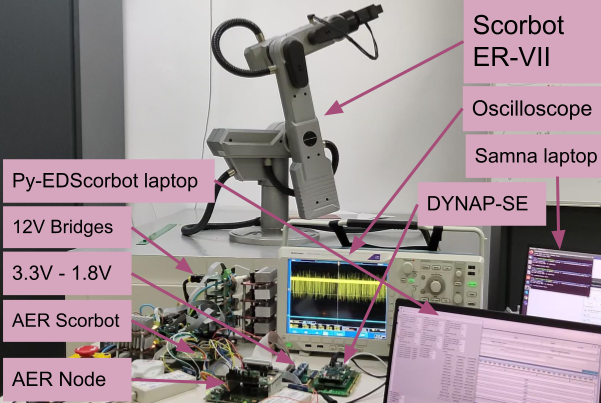}
    \subcaption{}
  \end{subfigure}
  \caption{(a) Block diagram of the ED-Scorbot with the DYNAP-SE chip interface. (b) Experiment set-up with the DYNAP-SE board connected to the ED-Scorbot framework.}
  \label{fig:SPID}
\end{figure*}

Therefore, there are several blocks needed to implement a communication between an external source and the effective driving of the robot's motors.
An overview of these blocks and their communication methods shown in Fig.~\ref{fig:scorbot_logical_overview}.
As the image depicts, the input signal can come from different sources.
This is due to the AER-Node programming.
For example, the current setup for the platform uses an external PC to configure and send this input.
However, we can tell it to accept other kinds of input, e.g. DYNAP-SE output spikes.
In order to do this, we have to take a closer look at the AER-Node board and its components to find an implementation that lets us take external spikes and integrate them into the communication.
The structure of the hardware for the SPID controllers and their interface through USB or DYNAP-SE is shown in Fig.~\ref{fig:SPID}.
As we can see, not all of the controller is located in the same board, due to resource scarcity.
The AER-Node board~\cite{aernode2017} and the AER-Scorbot~\cite{edscorbot2016} holds a Spartan 6 and a Spartan 3 FPGA respectively for the implementation of each part of the controllers.
The communication between both boards is through a 4-bit AER interface plus 2-bit for REQ and ACK handshake.
AER-Scorbot takes care of the final motor driving, expanding the pulses to be sent to the DC motors of the joints to allow the PFM signal of the SPID to power the motor. This, in fact, represents the proportional component of each SPID controller.
Furthermore, the optical encoder attached to each motor produces a phase shifted, two channel quadratic signal that codifies the rotation speed of the motor and the direction of the movement.
These signals are converted to spiking signals and codified into AER in the Spartan-3 FPGA.
In the other hand, the Spartan-6 FPGA implements the integrative and derivative part of the SPID controller, the feed-back of the controller with the conversion of the encoder information from speed to position (Integrate and Generate block of the feed-back), and the blocks for preparing the reference spiking signal for the SPID controllers, like the USB interface to spikes or the block developed in this work for interfacing the DYNAP-SE to the ED-Scorbot.
This interfacing block is explained with more details in next subsection.

\subsubsection{DYNAP-SE and ED-Scorbot interface}
The DYNAP-SE board provides 4 different parallel AER interfaces for sending and receiving its core neuron spikes in four different directions for configuring bigger spiking systems.
These ports are know as north, south, east and west.
% The python API connecting the PC to the chip, consists of an Opal-Kelly 7310 module connected to a set of one or many DYNAP-SE boards connected in a stacked way.
West port is used to send back spikes to the FPGA board, and then to the computer, the output spiking information produced by any spiking network configured on the chip. %DYNAP-SE stacked system.%
Nevertheless, the DYNAP-SE spiking output can be easily redirected through the east port of the DYNAP-SE chip.% in the stack, or the break-board at the bottom.

For connecting the DYNAP-SE chip to the robot, we are using the east port.
The East parallel AER bus of the chip has been connected to the AER input bus of the AER-Node board in the ED-Scorbot framework.
This connection has required the use of level converters (TXS0108E) to adapt the 3.3v of the GPIOs of the AER-Node to the 1.8V of the GPIOs of the DYNAP-SE.
As depicted in Fig.~\ref{fig:SPID}(a) top, the events produced by the chip should be processed by a sequence of blocks to reliably interpret the spiking output of the chip.
In the chip, a WTA network described in Section~\ref{subsec:dynpase} is producing a traffic for the wining population of neurons.
This WTA network output is configured to obtain as a winner an angle for one particular joint.
This winner ID has to be extracted from the two AER events output, so a finite-state-machine (FSM) is used for that in the AER Decoder block of the figure.
To avoid that the frequency of changes of the WTA can produce any vibration, high frequency changes or even damage in the robot, a history filter is included to produce a smooth sequence of changes in the robot's joints.
The filter is implemented by a integrate and fire neuron for each different input neuron ID, which is associated to each possible target angle of the joint, in such a way that the membrane potentials are incremented if an event is received from the population associated to that angle. The first neuron that reaches its threshold is selected to move the robot joint to the target angle. The selected angle neuron resets itself and sends an inhibitory spike to the other angles' neurons to reset their membrane potentials. The filtered neuron ID is the one that reaches the threshold firstly. Angle neurons' range and threshold values can be adjusted through the python interface \cite{pyedscorbot}.

%The filter is implemented by a set of a counter, a register and a comparator for each neuron ID.
%Each set acts as an integrate and fire neuron for each possible target angle of the joint, in such a way that the counters are incremented if an event is received from the population associated to that angle. The first population that reaches a threshold stored in the register is selected to move the robot joint to the target angle. The selected angle neuron resets itself and sends an inhibitory spike to the other angles' neurons to reset their counters.
%The filtered neuron ID is the one that reaches the threshold firstly. Counters' range and threshold value can be adjusted through the python interface \cite{pyedscorbot}.

Then, these filtered neuron IDs, that represent the next commanded angle for a joint, are mapped to the corresponding digital spike references values, which are used to feed a Spike Generator, and used as input for the SPID controllers.
Table \ref{tab:ref-angles} shows the neuron IDs used at the DYNAP-SE, their mapped angles, their digital spike references and their produced joint position values from the encoder's 16-bit internal counters.
\begin{table}[]
\caption{Correspondence for Neuron clusters, Joint angle, digital spiking reference and 16-bit Joint position (0º=Home)}
\label{tab:ref-angles}
\centering
\begin{tabular}{|c|c|c|c|}
\hline
\rowcolor[HTML]{EFEFEF} 
\multicolumn{1}{|l|}{\cellcolor[HTML]{EFEFEF}\textbf{Cluster (Neuron IDs)}} & \multicolumn{1}{l|}{\cellcolor[HTML]{EFEFEF}\textbf{Angle}} & \multicolumn{1}{l|}{\cellcolor[HTML]{EFEFEF}\textbf{Spike Ref}} & \multicolumn{1}{l|}{\cellcolor[HTML]{EFEFEF}\textbf{Position 16-bit}} \\ \hline
1(1-8)                                                               & 0º                                                          & 0                                                               & 32768                                                                 \\ \hline
\rowcolor[HTML]{EFEFEF} 
2(10-17)                                                             & 10,4º                                                        & 32                                                              & 34086                                                                 \\ \hline
3(19-26)                                                             & 20,8º                                                       & 64                                                              & 35406                                                                 \\ \hline
\rowcolor[HTML]{EFEFEF} 
4(28-35)                                                             & 31,2º                                                       & 96                                                              & 36724                                                                 \\ \hline
5(37-44)                                                             & 41,6º                                                       & 128                                                              & 38044                                                                 \\ \hline
\rowcolor[HTML]{EFEFEF} 
6(46-53)                                                             & 52,0º                                                       & 160                                                              & 39362                                                                 \\ \hline
7(55-62)                                                             & 62,4º                                                       & 192                                                              & 40682                                                                 \\ \hline
\rowcolor[HTML]{EFEFEF} 
8(64-71)                                                             & 72,8º                                                       & 224                                                             & 42000                                                                 \\ \hline
9(73-80)                                                             & 83,2º                                                       & 256                                                             & 43320                                                                 \\ \hline
\rowcolor[HTML]{EFEFEF} 
10(82-89)                                                             & 93,6º                                                       & 288                                                             & 44638                                                                 \\ \hline
11(91-98)                                                             & 104,0º                                                       & 320                                                             & 45958                                                                 \\ \hline
\rowcolor[HTML]{EFEFEF} 
12(100-107)                                                           & 114,4º                                                       & 352                                                             & 47276                                                                 \\ \hline
\end{tabular}
\end{table}
\subsection{Experiments}
\begin{figure}[ht] 
 \centering
 \includegraphics[scale=0.28]{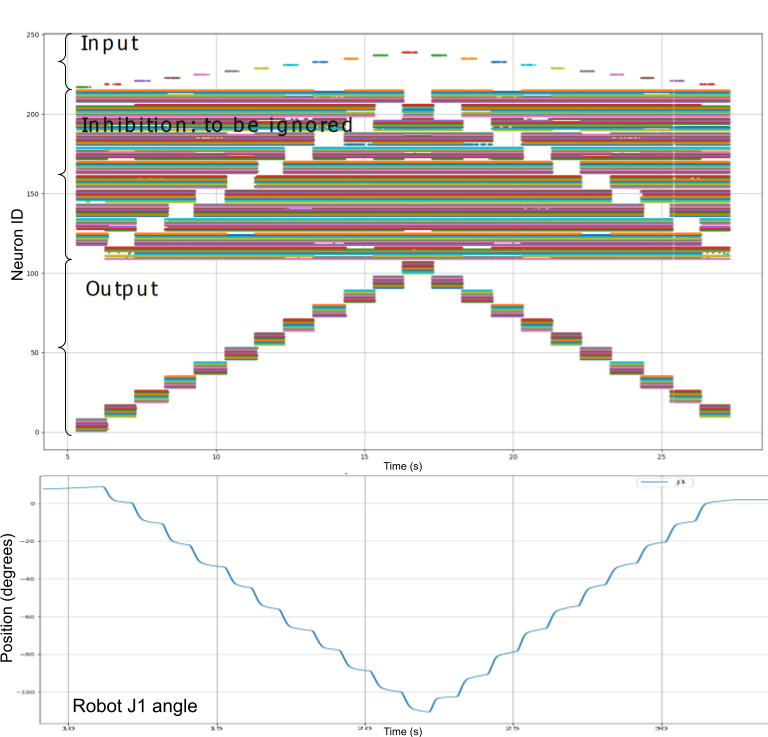}
 \caption{Proposed WTA network activity (top) with synthetic input, inhibitory activity and 12 populations output. Measured Joint 1 position response for this activity (bottom)}
 \label{fig:wta-activity}
\end{figure}

The proposed setup has been functionally validated with the execution of a SNN running at the DYNAP-SE that interacts with the base joint of the robot (joint 1). This SNN consists of a WTA network that selects one cluster over a set of 12 populations of neurons. Each population will command an angle to the joint according to Table \ref{tab:ref-angles}. The WTA has been stimulated programming its spike generator (i.e, input) to produce an incremental and decremental sequence of clusters activation. The joint is, therefore, changing its position accordingly. Figure \ref{fig:wta-activity} (top) shows the WTA input, its internal inhibitory spikes and its clusters activation over time. These output clusters activation neuronal activity is used by the AER Dyn2SPID block of Fig. \ref{fig:SPID}(a) for the joint control (see Fig. \ref{fig:wta-activity} bottom).

\section{Conclusions}
In this work a WTA circuit has been implemented over a neuromorphic spiking processor, i.e. DYNAP-SE, to emulate the behaviour of a WTA circuit of a biological CPG system for neuro-motor control. The output of this circuit has been connected to the ED-Scorbot robotic arm platform in real-time to perform a basic experiment to validate the circuit. 
It has been observed the impact of the temporal distribution of clusters activation in order to ensure the right target in the robot. SPID parameters \cite{EDScorbot-Springer}, motor torques and robot power supply imply a fixed latency for a joint to reach any commanded position. The larger the angle step among consecutive populations, the longer the required latency to reach the target position.  
Future work will be focused in combined pattern generation on several joints to reproduce learnt trajectories and close-loop WTA circuits.%in the exploration of the possibilities of close-loop WTA circuits when connected to the joints position signals.

%\section*{Acknowledgment}
% The preferred spelling of the word ``acknowledgment'' in America is without 
% an ``e'' after the ``g''.

% acknowledgments in the unnumbered footnote on the first page.

\bibliographystyle{IEEEtran}
\bibliography{biblioncs,IEEEabrv,IEEEexample, iscas_dynapse2edscorbot}
\vspace{12pt}
\end{document}